\documentclass{ieeeaccess}
\usepackage{cite}
\usepackage{amsmath,amssymb,amsfonts}
\usepackage{algorithmic}
\usepackage{graphicx}
\usepackage{textcomp}
\usepackage{multirow}
\usepackage{multicol}
\def\BibTeX{{\rm B\kern-.05em{\sc i\kern-.025em b}\kern-.08em
    T\kern-.1667em\lower.7ex\hbox{E}\kern-.125emX}}
\begin{document}
\history{Date of publication xxxx 00, 0000, date of current version xxxx 00, 0000.}
\doi{10.1109/ACCESS.2017.DOI} 

\title{Two-Dimensional Quantum Material Identification via Self-Attention and Soft-labeling in Deep Learning}
\author{
\uppercase{Xuan Bac Nguyen}\authorrefmark{1},
\uppercase{Apoorva Bisht}\authorrefmark{1}, 
\uppercase{Ben Thompson}\authorrefmark{1},
\uppercase{Hugh Churchill}\authorrefmark{2},
\uppercase{Khoa Luu}\authorrefmark{1},
\uppercase{Samee U. Khan}\authorrefmark{3}
}
\address[1]{Dept. of CSCE, University of Arkansas, Fayetteville, 72701, Arkansas, USA}
\address[2]{Dept. of Physics, University of Arkansas, Fayetteville, 72701, Arkansas, USA}
\address[3]{Dept. of Electrical and Computer Engineering, Mississippi State University, Starkville, 39762, Mississippi State, USA}

\markboth
{Author \headeretal: Preparation of Papers for IEEE TRANSACTIONS and JOURNALS}
{Author \headeretal: Preparation of Papers for IEEE TRANSACTIONS and JOURNALS}

\corresp{Corresponding authors: Khoa Luu (khoaluu@uark.edu)}

\begin{abstract}
In the quantum machine field, detecting two-dimensional (2D) materials in silicon chips is a critical problem. Instance segmentation is a potential approach to solve this problem. However, similar to other deep learning methods, instance segmentation requires a large-scale training data set and high-quality annotation in order to achieve considerable performance. Preparing the training data set is challenging in practice since annotators must deal with a large image, e.g., 2K resolution, and dense objects to annotate. In this work, we present a novel method to tackle the problem of missing annotation in instance segmentation in 2D quantum material identification. We propose a new mechanism for automatically detecting false negative objects, and we introduce an attention-based loss to the loss function to reduce the negative impact of these objects. We experiment on the 2D material detection data sets, and the experiments show that our method outperforms previous works.

\end{abstract}


\titlepgskip=-15pt

\maketitle

\section{Introduction}
\label{sec:introduction}
\PARstart{T}{wo}-dimensional (2D) materials hold immense potential for studying optical, electrical, thermal, and magnetic properties of materials and exploiting these properties at the micro-level. Till now, many physical methods \cite{Wang_2012} like Atomic Force Microscopy \cite{SINHARAY201374}, High-Resolution Transmission Electron Microscopy, Raman Spectroscopy \cite{ferrari2006raman}, and White Light Contrast Spectroscopy \cite{doi:10.1021/nl071254m} have been used to characterize the dimensions of 2D flakes. However, this is a tedious process, and the experimenter has to observe the flakes to determine their characteristics manually. It requires resources, skill, and time. Recently, deep-learning based approaches to identify and characterize 2D flakes have been proposed and implemented \cite{masubuchi2020deep, han2020deep}. Depending on the thickness of the flakes deposited on the dielectric layers, the optical images observed under the microscope show a gradient of colors. These colors are characteristic of the flake thickness and depend on the material used. For example, hBN (hexagonal boron nitride) flakes have a distinct color profile; similarly, graphene has its own color profile. It is this relationship among thickness, material, and color that the classification hinges on. The bottleneck is that an extensive data set is required to implement a fully functional and accurate deep-learning based application, and the data set consisting of images of flakes needs to be fully annotated. It means all image flakes should be identified, segmented, and annotated to train the model without false information.

In addition to identifying the thickness of the flake at specific pixels, the color gradient can be used to determine the quality of the flake due to changes in thickness. With more characteristics of 2D flakes identified based on microscope images, a deep-learning model can be trained to characterize not just the thickness but also the grade and other optical properties of the 2D flakes. It holds great promise to increase the efficiency and accuracy of the 2D flake "hunting" process.

To this end, we propose a novel method of instance segmentation that helps localize the location of 2D quantum materials using optical microscopy. We focus on solving the problem of missing annotations, resulting in lower recall and overall performance. Our contributions in this work can be summarized as follows,
\begin{itemize}
    \item We introduce a new mechanism based on attention for automatically detecting potential objects that do not have annotations. This method is trained end-to-end together with the full deep learning-based detection framework.
    \item We propose an attention-based loss strategy that can assign soft-labels for the objects detected above. This approach helps reduce the gradient impact of these objects contributing to the whole framework, thus increasing the recall and overall performance.
\end{itemize}

\begingroup
\newlength{\xfigwd}
\setlength{\xfigwd}{\textwidth}
\begin{figure*}[!t]
    \centering
    \includegraphics[width=1.50 \columnwidth]{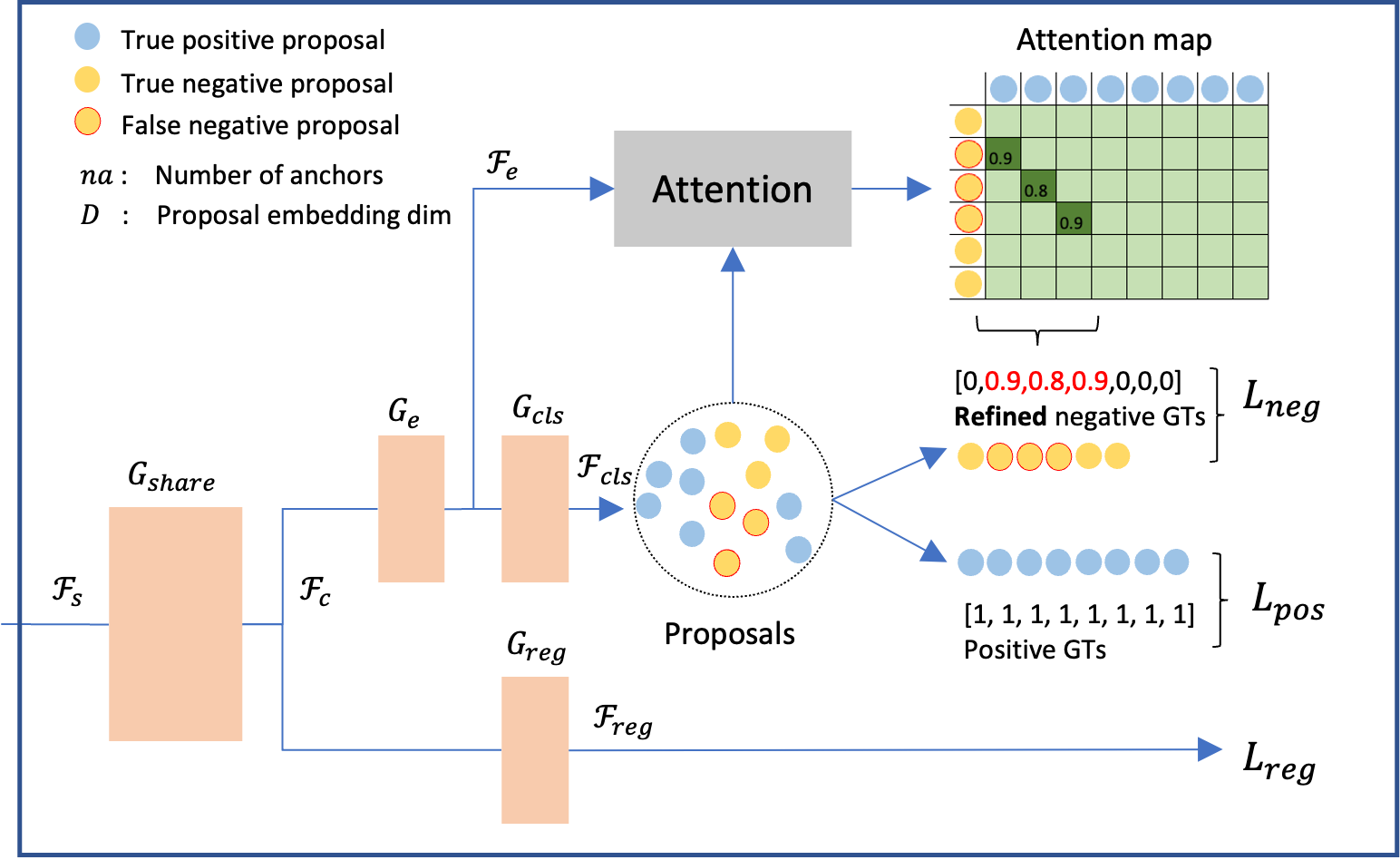}
    \caption{Our proposed Self-Attention and Soft-Labeling in Region Proposal Network.}
    \label{fig_1_rpn}
\end{figure*}
\endgroup

\section{Motivation}
\label{sec:motivation}
Until now, material scientists have been using various methods to determine the physical thickness. Some of these methods are summarized below. 

AFM (Atomic Force Microscopy): Although an AFM can provide resolutions as fine as a few fractions of a nanometer, it requires minute calibrations and has a limited scanning area of 150 by 150 micrometers. \cite{SINHARAY201374} It also has a slow throughput and can damage the crystal lattice during measurement. \cite{doi:10.1021/nl071254m}

TEM (Transmission Electron Microscopy): TEM utilizes a high-voltage beam of electrons to generate a highly magnified sample image. Properties of materials, like the thickness and defects, can be detected and imaged. However, the cost of a typical transmission electron microscope can be as high as \$10 million. \cite{winey2014conventional}

Raman Spectroscopy: Raman spectroscopy uses Raman signals from the materials to determine the thickness. Although a common method to determine the thickness of 2D materials, it has limitations. Raman spectroscopy gives the thickness of the materials at the location where the light is focused. Thus, many Raman measurements need to be taken to determine the thickness at multiple locations. It is time-consuming and often hard to map the entire flake. \cite{lyon1998raman}

White Light Contrast Spectroscopy: This method increases the contrast between the 2D flake and the substrate by utilizing the reflection spectrum from normal white light. \cite{yang2013rapid}

Among the techniques listed above, some have limitations, and most are expensive and require high maintenance. 

\section{Related Works}

The previous methods in this topic can be classified into two categories, including traditional and deep learning. They will be detailed in the following subsections. 

\subsection{Prior Works in 2D Quantum Material Identification}
\label{subsec:priorQM}

The classical methods for 2D Quantum Material Identification are that a researcher uses one of the methods listed in section \ref{sec:motivation} to obtain classification and thickness measurements. A common way to generate 2D flakes is by mechanical exfoliation. These exfoliated flakes are then transferred to a substrate like silicon and observed under the microscope. 

\subsection{Deep Learning in 2D Quantum Materials}
\label{subsec:priorDLQM}
2D quantum material research topics have gained attention from researchers in recent years. In particular, Bingnan et al. \cite{han2020deep} proposed a deep learning-based optical identification method to recognize 13 types of 2D materials from optical microscopic images. The method consists of several convolutions, batch norm \cite{ioffe2015batch}, ReLU \cite{agarap2018deep} followed by a softmax layer at the end. The method is inspired by UNet \cite{ronneberger2015u} architecture to segment the fakes with different thickness values. 
Satoru et al. \cite{masubuchi2020deep} presented an end-to-end pipeline for 2D flake identification. The method uses optical microscopy images as the input and outputs each flake's location and material type. The authors leverage Mask-RCNN \cite{he2017mask} to predict the bounding box and segmentation mask of different types of flake. This method achieves a breakthrough in the field of 2D quantum material detection.

\section{Data Collection}

In this work, we propose to develop an extensive training data set with thickness measurements of various flakes, ranging from around 5 nm to 100 nm, using a combination of optical microscopy and AFM. 
We start with silicon dioxide/silicon substrates already mounted with hexagonal boron nitride flakes. 
Since the substrate material and thickness alter the reflected light captured by the microscope, it was important to use consistent substrates. Switching substrates required recalibration to maintain the correspondence between color and thickness. hBN allows light to pass through it, and the reflected light from the substrate has a different spectrum based on the thickness of hBN. 

We use an optical microscope to capture these images. The images are calibrated using calibration slides so that images captured across different times of the day and different microscopes can be compiled for use in the same data set. 

We then use Atomic Force Microscopy to measure the thickness of each flake captured via optical microscopy. Thus, we get color (RGB) information via the microscope image and thickness information via the AFM. It can be combined to create a data set that establishes color-to-thickness relation. 
To do this, we create two types of data sets: 
\begin{enumerate}
    \item In the first version, we search for uniform-looking flakes and measure their thickness via the AFM. Since the variation in thickness is estimated to be below the target error margin of 5 nm, an average thickness is determined from various random spots on the flake, and a thickness value is assigned to it. Flake thickness is then annotated in the 20x optical microscope images. It means that every selected flake in the optical microscope image has associated metadata referring to the average thickness of the flake. 
    \item In the second version, we aim to generate aligned AFM and optical microscope images of individual flakes. The two images perfectly superimpose each other, and information from one image serves as metadata for the second image (thickness information from AFM and RGB values from optical microscope images). It helps us discard the bias towards uniform thickness flakes which means that the dataset can accommodate a much larger diversity of thickness information. However, we need higher-resolution AFM and microscope images to enable this superposition between each pixel (or group of pixels). Thus, in version \#2, we use $256*256$ resolution AFM scans (over a region spanning 20$\mu m^2$ to 40$\mu m^2$) and 50x optical microscope images. 
\end{enumerate}

The following paragraphs describe some considerations considered during data collection and model training. 

\paragraph{Substrate thickness}
Our current data set uses a uniform-thickness silicon dioxide (SiO$_2$) substrate. However, different applications and groups use varying thicknesses, which results in variations in the apparent color of the 2D material. Since the model recognizes the thickness based on color, such drivers (like substrate thickness) can change the wavelength of the light reflected to create a bottleneck. Thus, we need proper calibration for every configuration.
\paragraph{Microscope calibration}
Different research groups utilize microscopes, light sources, exposure times, RGB, and white balance. It means the same flake can have different apparent colors under different microscopes. It, again, poses an issue and is a bottleneck to achieving fairness. We use the same calibration across all samples and images to develop our data set. To make the application available to multiple research institutes, a one-time built-in calibration step should allow users to calibrate the application once and use it for their purpose. 
\paragraph{Materials}
A wide variety of 2D materials are utilized in various research labs, and all of them have different thickness-dependent color spectrums based on their lattice structure. It means that training the model for one material implies that the application can be reliably used for that model. Thus, so far, in order to use the application for multiple materials, the model needs to be separately trained. 
\paragraph{Dataset} 
We collected 101 samples of hBN and 30 samples of Graphene, including their annotated pictures on the optical microscope and their thickness measurements on the AFM. They were labeled in (enter labeling software). The process of collecting these samples was both tedious and arduous. 

First, the optical microscope had to be calibrated, and we had to save a calibration picture. Then, the chip was loaded into the optical microscope, and we had to hunt for flakes to measure the thickness. After taking individual pictures of each flake, we took a stitched picture of all flakes to determine the chip's position. 

We then loaded the chip into the AFM. After calibrating many parts of the AFM, including the (voltage?) and (Z-value?), we could hunt for the same flakes we imaged in the optical microscope. After finding a flake, we had to approach the flake with the microscopic tip of the AFM that vibrates over the sample and collects the thickness data. The AFM then begins measuring the flake's thickness, and even for small flakes, this takes at least 30 minutes. 

Upon obtaining this thickness measurement, we have to use software to remove noise from the image and make it cleaner. Finally, we have to annotate the image from the optical microscope. A common issue is a broken or blunted tip in the AFM, which results in corrupted and sometimes nonsensical thickness measurements. The tip must then be replaced, which is another tedious task.

\section{The Proposed Method}
Our proposed method employs Mask-RCNN \cite{he2017mask} as the baseline architecture. It contains three main components: the Backbone ($B$), the Region Proposal Network ($RPN$), and the Region of Interest Head ($ROI$).

Let $I \in \mathbb{R}^{H \times W \times C}$ be the input image, where $H, W, C$ are height, width, and number of channels correspondingly. The backbone $B$ extracts deep feature maps of $I$ denoted as $\mathcal{F}_s = B(I), \mathcal{F}_s \in \mathbb{R}^{H_s \times W_s \times C_s}$ where $s$ is scale and $H_s = H / s$ and $W_s = W / s$.

Mask-RCNN is a two-stage framework where the results of the first stage are passed as input to the second stage. If the first stage is not good enough, the Performance of the second stage will fall apart. The first stage includes $RPN$ that also nominates the potential objects and their shapes represented by anchors. 
Apart from prior methods \cite{girshick2015fast} using selective search, $RPN$ can be learned via back-propagation. In the case of missing annotation, some objects do not have any bounding boxes. Typically, $RPN$ is confused by treating these potential objects as a background while some similar objects are considered as a foreground. In other words, missing annotation is a form of existing \textit{false negative} and leads to \textit{lower recall} of proposing potential objects. As a result, the second stage will be influenced. 

In this paper, our main contribution is to \textit{focus on the $RPN$ to manipulate the missing annotation problem}, enhance the confidence of proposed objects, and subsequently improve the next stage and overall Performance. Our proposed method is illustrated in Fig. \ref{fig_1_rpn}.

In the first stage, $RPN$ receives $\mathcal{F}_s$ as an input and outputs a list of object proposals $\mathcal{P}_s$.
\begin{equation}
\label{rpn_in_out}
    \mathcal{P}_s = RPN(\mathcal{F}_s)
\end{equation}
Let $na$ be the number of anchors, then the total number of proposals that can be generated is $|\mathcal{P}_s| = H_i \times W_i \times na$. $\mathcal{P}_s$ is split into two subsets: $\mathcal{P}_{pos}$ for positive proposals and $\mathcal{P}_{neg}$ for negative proposals.
\begin{align}
    \mathcal{P}_s &= \mathcal{P}_{pos} + \mathcal{P}_{neg}, \; |\mathcal{P}_s| = H_i \times W_i \times na = |\mathcal{P}_{pos}| + |\mathcal{P}_{neg}|
\end{align}
The objectives of $RPN$ are: \textit{(1).  Predicting if a proposal is foreground (1) or background (0)} and \textit{(2). Estimating anchor delta}.
\begin{align}
\label{eq_rpn_losses}
\mathcal{L}_{RPN} &= \mathcal{L}_{cls} + \mathcal{L}_{reg} \\ \nonumber
                  &= \frac{1}{|\mathcal{P}_s|}\sum_{p_i \in \mathcal{P}_s} L_{cls}(p_i, \hat{p_i}) + \frac{1}{|\mathcal{P}_{neg}|}\sum_{p_i \in \mathcal{P}_{neg}} L_{reg}(t_i, \hat{t_i})
\end{align}
where $ L_{cls}$ and $ L_{reg}$ are the loss functions. $p_i$ is the probability the anchor $i^{th}$ is foreground while $\hat{p_i}$ is the ground truth.  Similarly, $t_i, \hat{t_i}$ are predictions and ground truth of anchor size \cite{he2017mask} respectively. It is noted that $\mathcal{P}_s$ includes both negative and positive proposals. Thus, in the Eq. \eqref{eq_rpn_losses}, $\mathcal{L}_{reg}$ encounters the positive proposals only, while $\mathcal{L}_{cls}$ involves all type of proposals. The $\mathcal{L}_{cls}$ can be reformulated as follows,
\begin{equation}
\small
\begin{split}
     \mathcal{L}_{cls} &= \mathcal{L}_{pos} + \mathcal{L}_{neg} \nonumber \\
                      &= \frac{1}{|\mathcal{P}_{pos}|}\sum_{p_i \in \mathcal{P}_{pos}} L_{cls}(p_i, \hat{p_i}) \nonumber + \frac{1}{|\mathcal{P}_{neg}|}\sum_{p_i \in \mathcal{P}_{neg}} L_{cls}(p_j, \hat{p_j})
\end{split}
\end{equation}
It is clear that $\forall p_i \in \mathcal{P}_{pos}, \hat{p_i} = 1$ and $\forall p_j \in \mathcal{P}_{neg}, \hat{p_j} = 0$. However, dual to missing annotations, it leads to $\exists p_j \in \mathcal{P}_{neg}$ such that $\hat{p_j} = 1$ and $p_j$ is called as \textit{false negative} samples. We propose a mechanism to detect false negative samples automatically. Besides, we present an alternative loss function that uses a soft label for these samples to help reduce gradient impacts on the overall architecture.
As in Eq. \eqref{rpn_in_out}, $\mathcal{F}_s$ is the input, $RPN$ is designed as follows,

\begin{align}
    \mathcal{F}_c &= G_{share}(\mathcal{F}_s)\;\;\;\; \mathcal{F}_{reg} = G_{reg}(\mathcal{F}_c) \nonumber \\
    \mathcal{F}_{e} &= G_{e}(\mathcal{F}_c)\;\;\;\;\;\;\;\;\;\;\; \mathcal{F}_{cls} = G_{cls}(\mathcal{F}_e)
\end{align}
First, we use a convolution layer $G_{share}$ to create a $\mathcal{F}_c$ feature map. It is used to estimate anchor delta values of each anchor by passing to a convolution $G_{reg}$. Besides, $\mathcal{F}_c$ is also used to create embedding features $\mathcal{F}_e$ for each anchor using $G_e$, where $G_e$ denotes the embedding feature.

\begin{figure}[!t]
    \centering
    \includegraphics[width=1.0\columnwidth]{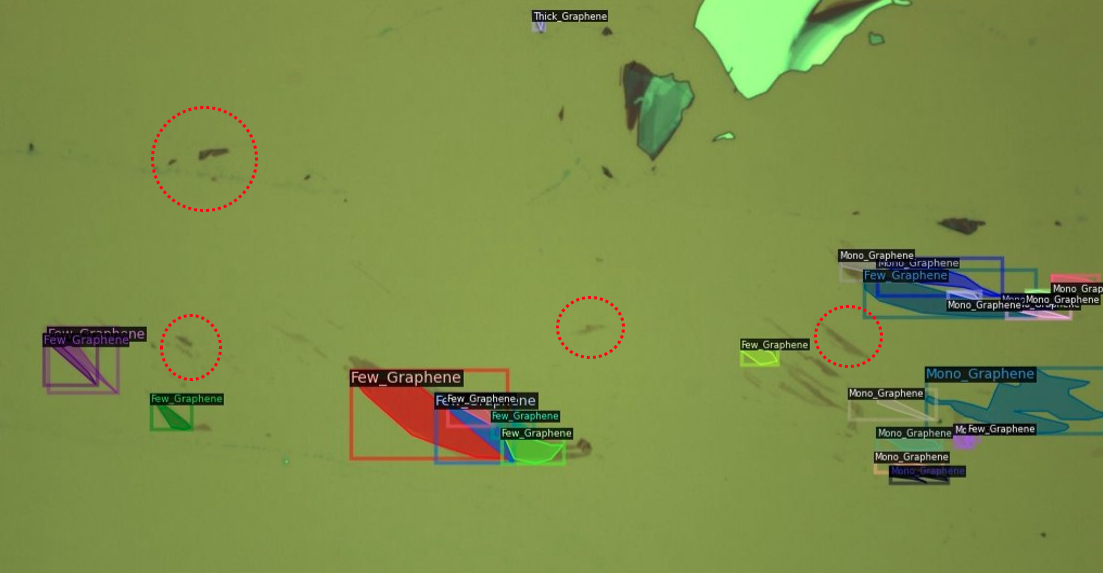}
    \vspace{-2em}
    \caption{The sample of Graphene objects. The red dot circles represent a false negative object without any annotations.}
    \vspace{-1em}
    \label{fig:dataset_example}
\end{figure}

It is followed by $G_e$ is  $G_{cls}$, which is used to estimate the probability of every anchor belonging to the foreground or background.
It is noted that $\mathcal{F}_e \in \mathbb{R}^{H_e \times W_e \times (na \times D)}$ and $\mathcal{F}_{cls} \in \mathbb{R}^{H_{cls} \times W_{cls} \times na}$, where $D$ is the dimension of embedding feature and $na$ is the number of anchors. $G_{cls}$ is designed as a convolution with kernel size and stride as 1 to make sure that: $H_e = W_{cls}, H_w = H_{cls}$ and every anchor has its corresponding embedding.

The proposal set $\mathcal{P}_s$ is generated from $\mathcal{F}_{cls}$, and proposals can be identified as either positive or negative ones. We measure the self-attention between two sets of proposal $\mathcal{P}_{pos}$ and $\mathcal{P}_{neg}$ as follows,
\begin{align}
    A = softmax(Norm(\mathcal{F}_{neg}) \odot Norm(\mathcal{F}_{pos}))
\end{align}

where $\mathcal{F}_{pos} \in \mathbb{R}^{N_{pos} \times D}$, $\mathcal{F}_{neg} \in \mathbb{R}^{N_{neg} \times D}$ are feature set of all positive and negative proposals. $Norm$ is denoted as the normalization function and $\odot$ is matrix multiplication. $A \in \mathbb{R}^{N_{neg} \times N_{pos}}$ is the attention map, $A_{ij}$ represents for the similar score between $i^{th}$ negatives and $j^{th}$ positive sample and $\sum_{j}A_{ij} = 1$. We expect a high $A_{ij}$ score for the false negative proposals and a low score for the true negative one. We define the threshold $t$ to determine which sample is false negative if $A_{ij} > t$. The loss function for negative proposals is reformulated as follows.

\begin{align*}
    L_{neg}(p_i, \hat{p}_i) = \begin{cases}
                            	L_{cls}(p_i, max(A_i)), \text{ if } max(A_i) \ge t \\
                            	L_{cls}(p_i, 0), \text{ if } max(A_i) < t
                            \end{cases}
\end{align*}

\section{Experiments}
\subsection{Quantum Image Datasets}
We use the quantum image dataset published in \cite{masubuchi2020deep}. The dataset contains 4 type of materials: hBN, Graphene, MoS2, and WTe2. The number of corresponding images for each material are: 353, 862, 569, and 318 images and the number of objects for each material types are: 456, 4805, 839, and 1053 respectively. Fig. \ref{fig:dataset_example} shows a sample from dataset.

\subsection{Experiment Settings}
In our experiments, we use detectron2 framework that is based on Pytorch. We follow the protocol and use the same hyper parameters as described in \cite{zhang2020solving}. More specially, we employ Resnet-101 \cite{he2016deep} as the backbone and FPN \cite{lin2017feature} for multilevel features enhancement. The number of anchors $na=3$, the feature dimension of anchor's embedding is $D=256$. We train the network with learning rate as 0.02 in 10000 iterations and gradually decrease 10 times at the milestone of 5000 and 8000 iterations. The bach size is set to 48. The model is trained on a machine of 4GPUs with 48G each. Training time takes 2 hours to finish. We understand that the baseline in \cite{masubuchi2020deep} uses Keras/Tensorflow implementation, so the model is retrained by using detectron2 as a baseline.

\subsection{Thickness Estimation}

Along with the flake detection, we also build a model to estimate the thickness of the detected flakes. This module is a linear regression machine learning algorithm implemented with Sklearn. While this model is intended to be used on flakes that are uniform in color/thickness, it can also be used on less consistent flakes to predict the average thickness. 

This module takes as input the image masks as predicted by the detection module. For each mask, \textit{n} crops are taken randomly from inside the mask. The crops then underwent a series of transformations, i.e., horizontal flips, vertical flips, and $90^\circ$ rotations, each with a probability of 50\%

Each crop, \textit{C}, was then flattened from an RGB image into a 1D array, \textit{x}. 
\begin{equation}
    C \in \mathbb{R}^{H\times W \times C} \to x \in \mathbb{R}
\end{equation}
and the input array \textit{X} was constructed by appending all flattened crops
\begin{equation}
    X = \{x_i\}_{i=0}^{i=nk -1}
\end{equation}
where \textit{k} is the number of flakes in the training set. The input array \textit{y} is constructed by appending the thickness of each crop such that each thickness is included \textit{n} times. It is demonstrated in Equation \ref{eqn:y}, where $T_{i}$ represents the thickness of the flattened crop $x_{i}$. 
\begin{equation}
    \label{eqn:y}
    y = \{T_{i}\}_{i=0}^{nk-1} 
\end{equation}

The inputs \textit{X} and \textit{y} are then passed into the regression algorithm. This algorithm plots the dependent variable (thickness) against the independent variable (color) to find the linear function that best represents the relationship between the two. 

Using this function, the algorithm predicts the thickness for each pixel within a crop. Each of these predictions is averaged together to get the thickness of that crop. This step is repeated \textit{n} times for each of the \textit{n} crops, and each of these is averaged again for the final thickness prediction of the flake.

\begin{table}
    \centering
    \begin{tabular}{|l|l|c|}
    \hline 
         Train & Test & Error \\ 
         \hline 
         hBN & hBN* & $4.0$ nm \\ 
         hBN & Graphine & $8.15$ nm \\ 
         Graphine & Graphine* & $7.39$ nm \\ 
         hBN \& Graphine & hBN \& Graphine* & $5.6$ nm \\ 
    \hline 
    \end{tabular}
    \caption{Performance on different types of 2D quantum materials using different training data. An asterisk denotes the use of $k$-fold cross-validation.}
    \label{tab:thickness_results}
\end{table}

\begin{table}[ht!]
\centering
\resizebox{.5\textwidth}{!}{
\begin{tabular}{|l|l|c|c|c|c|c|c|}
    \hline
    \multirow{2}{*}{Type} & \multirow{2}{*}{Methods} & \multicolumn{3}{c|}{Bounding Box} & \multicolumn{3}{c|}{Segmentation} \\
    \cline{3-8}
    & & \bf AP$_{\bf 50}$ & \bf AP$_{\bf 75}$ & \bf AP & \bf AP$_{\bf 50}$ & \bf AP$_{\bf 75}$ & \bf AP \\
    \hline
    \multirow{6}{*}{G} & Baseline & 86.3 & 78.8 & 72.4 & 84.1 & 71.2 & 63.2 \\
                            & OHEM \cite{shrivastava2016training} & 86.5 & 79.8 & 73.1 & 84.5 & 71.4 & 63.7 \\
                            & BRL \cite{zhang2020solving} & 86.8 & 79.8 & 73.6 & 85.1 & 71.5 & 63.8 \\
    \cline{2-8}
    & Ours ($t=0.6$) & 87.1 & 80.3 & 74.6 & 85.0 & 72.7 & 64.6 \\
    & Ours ($t=0.8$) & \textbf{87.8} & \textbf{80.8} & \textbf{75.7} & \textbf{85.7} & \textbf{73.8} & \textbf{65.4} \\
    & Ours ($t=0.9$) & 87.1 & 80.2 & 74.1 & 84.5 & 72.4 & 64.3 \\
    \hline
    
    \multirow{6}{*}{BN} & Baseline & {73.2} & 62.0 & 59.0 & 73.4 & 73.2 & {62.9} \\
                         & OHEM \cite{shrivastava2016training} & 71.4 & 71.0 & 61.8 & 71.4 & 62.0 & 61.4 \\
                         & BRL \cite{zhang2020solving} & 71.6 & 71.4 & 62.8 & 71.0 & 69.8 & 60.8 \\
    \cline{2-8}
                         & Ours ($t=0.6$) & 72.1 & {72.1} & {63.7} & {72.1} & {71.3} & 62.7 \\
                         & Ours ($t=0.8$) & \textbf{73.4} & \textbf{73.1} & \textbf{64.4} & \textbf{73.1} & \textbf{72.9} & \textbf{63.0} \\
                         & Ours ($t=0.9$) & 70.3 & 70.3 & 62.1 & 70.3 & 70.3 & 60.3 \\
    \hline
    \multirow{6}{*}{M} & Baseline & 60.8 & 57.0 & 50.3 & 60.3 & 54.3 & 47.6 \\
                          & OHEM \cite{shrivastava2016training} & 62.5 & 56.1 & 51.1 & 61.7 & 54.1 & 48.4 \\
                          & BRL \cite{zhang2020solving} & 63.7 & 59.4 & 53.7 & 63.0  & 57.0 & 50.6 \\
    \cline{2-8}
                          & Ours ($t=0.6$) & 64.0 & {57.8} & {54.2} & {63.3} & {57.5} & 50.9 \\
                          & Ours ($t=0.8$) & \textbf{65.6} & \textbf{60.3} & \textbf{55.1} & \textbf{64.9} & \textbf{58.1} & \textbf{51.9} \\
                          & Ours ($t=0.9$) & 64.1 & 57.9 & 52.0 & 63.9 & 60.2 & 51.8 \\
    \hline
    \multirow{6}{*}{W} & Baseline & 76.6 & 64.8 & 58.6 & 74.2 & 60.7 & 52.3 \\
                          & OHEM \cite{shrivastava2016training} & 76.7 & 65.3 & 59.8 & 75.5 & 61.2 & 54.4 \\
                          & BRL \cite{zhang2020solving} & \textbf{79.6} & 66.0 & 60.6 & \textbf{78.0} & 61.4 & 55.0 \\
    \cline{2-8}
                          & Ours ($t=0.6$) & 78.3 & 65.3 & 61.4 & 77.0 & {64.6} & \textbf{55.9} \\
                          & Ours ($t=0.8$) & {78.8} & {67.6} & \textbf{61.8} & {77.3} & {63.2} & 55.3 \\
                          & Ours ($t=0.9$) & 79.0 & \textbf{68.5} & 61.2 & {77.4} & \textbf{65.0} & 55.4 \\
    \hline
\end{tabular}
}
\vspace{-0.6em}
\caption{Performance on different types of 2D quantum materials. 
G: \textit{Graphene}, BN: \textit{hBN}, M: \textit{MoS2}, W: \textit{WeT2}}
\vspace{-0.5em}
\label{tab:performance_2d_materials}
\end{table}

\section{Results}
The performance on four types of materials is reported in Table \ref{tab:performance_2d_materials} following official COCO metrics \cite{lin2014microsoft}. Overall, our method outperforms previous methods \cite{masubuchi2020deep}, \cite{zhang2020solving} in terms of average precision of both bounding box and segmentation criteria. More specially, compared to the baseline \cite{masubuchi2020deep}, there are large margins of improvement: approximately 3\% - 5\% in bounding boxes and 3\% of in segmentation criteria. Besides, our method outperforms BRL \cite{zhang2020solving} in a different type of materials and average precision by approximately 2\%. 

\begin{figure}[!t]
    \centering
    \includegraphics[width=1.0\columnwidth]{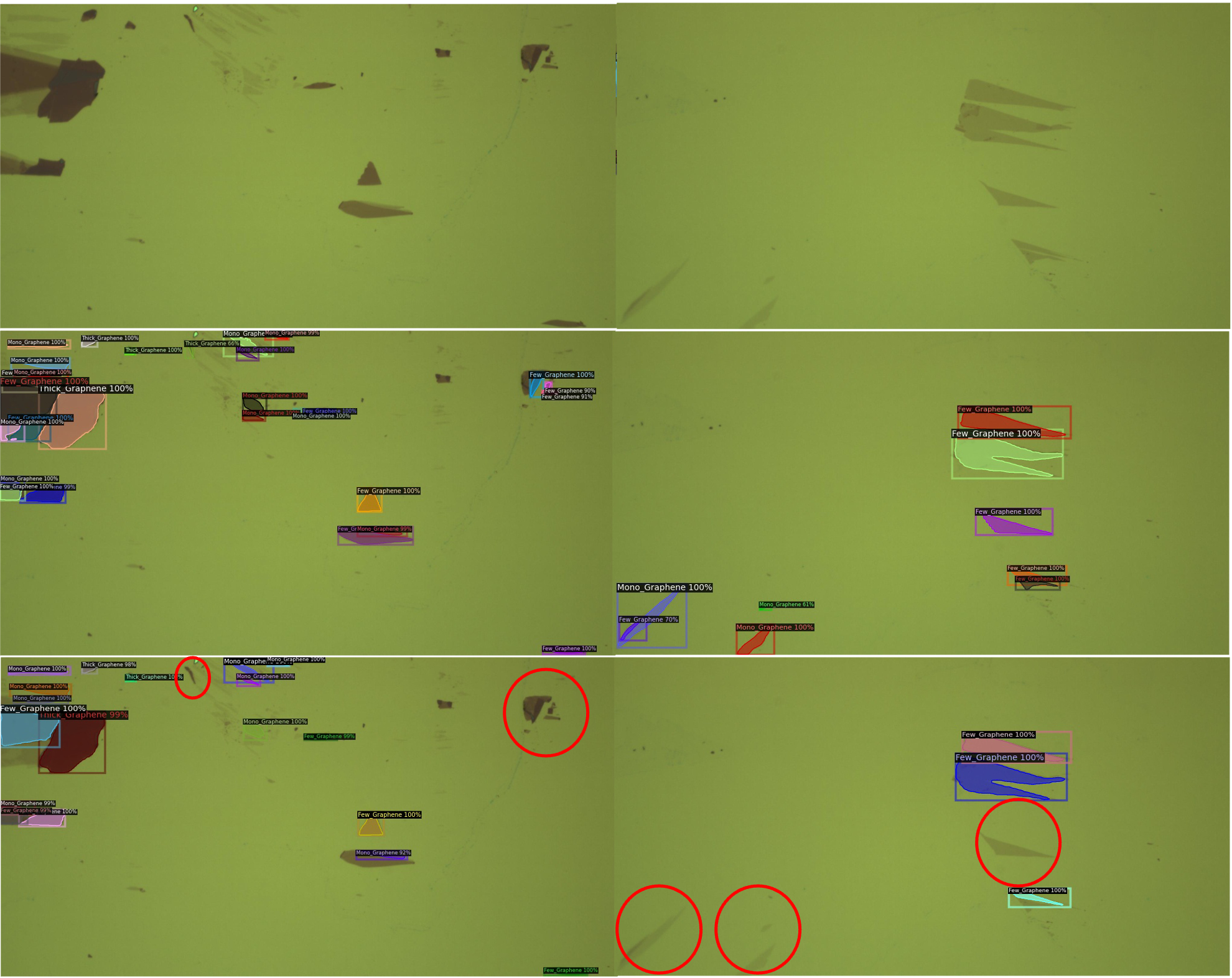}
    \caption{Qualitative results. The first row is the input images, the second row is our results, and the third row is results of baseline.}
    \label{fig:quality_results}
\end{figure}

In addition, we implement the ablation experiments with different threshold values $t=0.6, 0.8, 0.9$ to investigate how the Performance varies. We found that threshold $t = 0.6$ is enough to surpass previous methods, and $t = 0.8$ is the optimal value. The qualitative results are shown in Fig. \ref{fig:quality_results}. 

The performance in predicting the thickness of two materials is reported in Table \ref{tab:thickness_results}. As expected, the error in our thickness results is the lowest when the testing material matches the training material. We achieve a $4.0$ nm error in the best run when our training and testing material is hBN. 

\section{Conclusions}
This work has presented a new approach to tackling the missing annotation problem in instance segmentation in 2D quantum material identification. A new mechanism has been introduced to detect false negative objects and reduce their negative impact with an attention-based loss strategy. The experimental results on 2D material datasets have shown the advantages of our work compared to prior work by a large margin.

\bibliographystyle{unsrt}
\bibliography{sn-bibliography}

\begin{IEEEbiography}[{\includegraphics[width=1in,height=1.25in,clip,keepaspectratio]{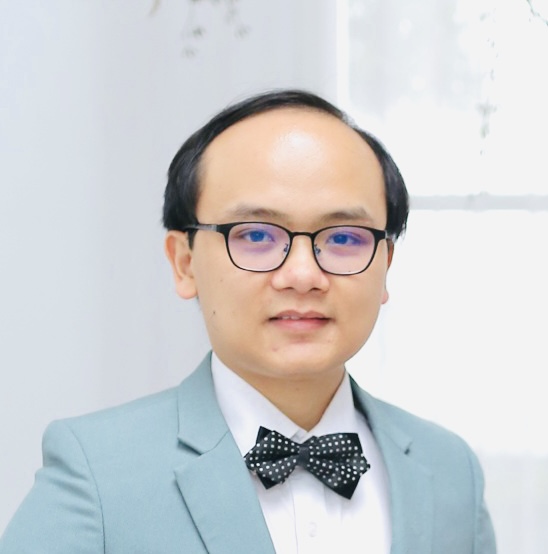}}]{Xuan-Bac Nguyen}
is currently a Ph.D. student at the Department of Computer Science and Computer Engineering of the University of Arkansas. He received his M.Sc. degree in Computer Science from Electrical and Computer Engineering department, Chonnam National University, South Korea, 2020.  He received his B.Sc. degree in Electronics and Telecommunications, University of Engineering and Technology, VNU in 2015. In 2016, he was an software engineer at Yokohama, Japan. His research interests include Face Recognition, Facial Expression, and Medical Image Processing.
\end{IEEEbiography}

\begin{IEEEbiography}
[{\includegraphics[width=1in,height=1.25in,clip,keepaspectratio]{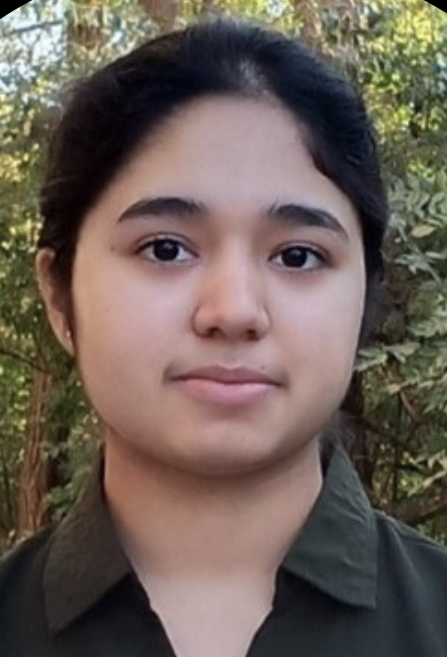}}]{Apoorva Bisht} received her Bachelor of Science degrees in physics (honors) and computer science (honors) from University of Arkansas in 2023. As an undergraduate research assistant in labs in physics and computer science department, her research projects dealt with non-linear and quantum optics, and using AI to characterize 2D materials. She will be pursuing her PhD at University of Colorado, Boulder in AMO physics.

\end{IEEEbiography}

\begin{IEEEbiography}
[{\includegraphics[width=1in,height=1.25in,clip,keepaspectratio]{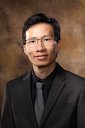}}]{Khoa Luu} (Member, IEEE) was a Research 
Project Director with the Cylab Biometrics Center, 
Carnegie Mellon University (CMU), USA. He is 
currently an Assistant Professor and the Director 
of the Computer Vision and Image Understand- 
ing (CVIU) Laboratory, Department of Computer 
Science and Computer Engineering, University of 
Arkansas, Fayetteville, USA. He has received four 
patents. He has coauthored more than 120 papers 
in conferences and journals. His research inter- 
ests include face recognition, biometrics, video and image understanding, 
autonomous car driving, robot vision, computer vision, machine learning, 
and quantum machine learning. 
Dr. Luu is a PC Member of AAAI and ICPRAI, in 2020 and 2022, 
respectively. He received two best paper awards. He was a Vice-Chair of 
Montreal Chapter IEEE SMCS, Canada, from September 2009 to March 
2011. He is a Co-organizer and the Chair of CVPR Precognition Workshop, 
in 2019, 2020, 2021, and 2022; MICCAI Workshop, in 2019, 2020; and 
ICCV Workshop, in 2021. He is serving as an Associate Editor of IEEE 
ACCESS journal. He is currently a Reviewer for several top-tier conferences 
and journals, such as CVPR, ICCV, ECCV, NeurIPS, ICLR, FG, BTAS, 
IEEE TRANSACTIONS ON PATTERN ANALYSIS AND MACHINE INTELLIGENCE, IEEE 
TRANSACTIONS ON IMAGE PROCESSING, Journal of Pattern Recognition, Journal 
of Image and Vision Computing, Journal of Signal Processing, Journal of 
Intelligence Review, IEEE ACCESS, and IEEE TRANSACTIONS. 

\end{IEEEbiography}

\begin{IEEEbiography}
[{\includegraphics[width=1in,height=1.25in,clip,keepaspectratio]{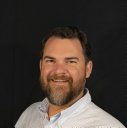}}]{Hugh Churchill} was a Pappalardo postdoctoral fellow at MIT. He is currently an Assistant Professor of Physics and Principal Investigator at the Churchill Lab in the Department of Physics at the University of Arkansas, Fayetteville, USA. He has coauthored 91 papers. His research interests include condensed matter physics, quantum materials and devices, optoelectronics, quantum dots, two dimensional materials, and microfabrication. 

\end{IEEEbiography}

\begin{IEEEbiography}
[{\includegraphics[width=1in,height=1.25in,clip,keepaspectratio]{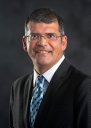}}]{Samee Ullah Khan} (Senior Member, IEEE) received a Ph.D. in 2007 from the University of
Texas. He is the Department Head and James
W. Bagley Chair Professor of Electrical \& Computer Engineering at Mississippi State University
(MSU). Before arriving at MSU, he was Cluster
Lead (2016-2020) for Computer Systems Research at National Science Foundation and the
Walter B. Booth Professor at North Dakota State
University. His research interests include optimization, robustness, and security of computer
systems. His work has appeared in over 450 publications. He is the
associate editor of IEEE Transactions on Cloud Computing and the
Journal of Parallel and Distributed Computing.
\end{IEEEbiography}

\begin{IEEEbiography}
[{\includegraphics[width=1in,height=1.25in,clip,keepaspectratio]{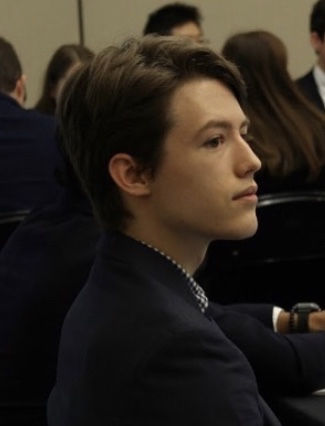}}]{Benjamin Thompson} is an undergraduate researcher at the University of Arkansas pursuing degrees in Mathematics and Computer Science. He is an undergraduate research assistant in the Computer Vision and Image Understanding laboratory at the University of Arkansas. His research interests include computer vision, machine learning, quantum computing, and optimization.
\end{IEEEbiography}

\EOD

\end{document}